\documentclass{article} 
\usepackage{iclr2027_conference,times}

\usepackage{amsmath,amsfonts,bm}

\def\eqref#1{equation~\ref{#1}}

\def\1{\bm{1}}

\DeclareMathAlphabet{\mathsfit}{\encodingdefault}{\sfdefault}{m}{sl}
\SetMathAlphabet{\mathsfit}{bold}{\encodingdefault}{\sfdefault}{bx}{n}

\usepackage{hyperref}
\usepackage{url}
\usepackage{booktabs}
\usepackage{float}
\usepackage{graphicx}
\usepackage{wrapfig}
\usepackage{subcaption}
\usepackage{fancyvrb}
\title{
OPMD: On-Policy Mechanism Distillation\\
Beyond Output Supervision
}

\author{Yingdong Shi, Zhihong Li, Changming Li, Jingyi Yu\textsuperscript{$\dagger$}, Quan Li\textsuperscript{$\dagger$}, Kan Ren\textsuperscript{$\dagger$} \\
ShanghaiTech University \\
\texttt{\{shiyd2026,yujingyi,liquan,renkan\}@shanghaitech.edu.cn}
}

\hypersetup{
    pdftitle={OPMD: On-Policy Mechanism Distillation Beyond Output Supervision},
    pdfauthor={Yingdong Shi, Zhihong Li, Changming Li, Jingyi Yu, Quan Li, Kan Ren}
}

\newcommand{\method}{\textsc{OPMD}}

\iclrfinalcopy 
\begin{document}

\maketitle
\lhead{Preprint}
\begingroup
{
\renewcommand{\thefootnote}{\fnsymbol{footnote}}
\footnotetext[2]{Correspondence to Jingyi Yu, Quan Li, and Kan Ren.}
}
\endgroup

\begin{abstract}

On-policy distillation (OPD) improves language-model by supervising a student on prefixes sampled from its own policy, but its supervision remains confined to the output space.
Consequently, OPD transfers what a teacher predicts without supervising the intermediate computation that produces those predictions.
Simply distilling intermediate representations like representation matching is insufficient since teacher and student representations may differ geometrically, and even well-aligned hidden states need not be usable by the student's subsequent computation.
We introduce On-Policy Mechanism Distillation (\method), which transfers teacher-derived intermediate computation while explicitly accounting for its compatibility with the student.
OPMD first constructs a student-compatible teacher that transforms teacher representations into intermediate states usable by the student's downstream computation, then it internalizes these states from on-policy trajectories. 
OPMD achieves the highest average accuracy among the evaluated distillation baselines in all three teacher--student settings, and generalizes to multi-teacher distillation and cross-family supervised fine-tuning.
Further analyses show that mechanism supervision yields more effective reasoning and meaningfully alters the student's intermediate computation and solution strategies, supporting that it transfers information beyond output supervision alone.
Together, these results establish mechanism-level supervision as an effective way to move on-policy distillation beyond output supervision, enabling students to learn inner mechanisms they can use and internalize.
\end{abstract}

\section{Introduction}
On-policy distillation supervises a student on textual prefixes sampled from its own policy, addressing a key mismatch between training and deployment \citep{agarwal2024onpolicy,gu2024minillm}.
However, agreement on observed next-token distributions does not establish agreement about how evidence is represented and transformed across layers.
Models with similar next-token distributions can organize evidence differently in the representation space and rely on different computational pathways at depth~\citep{belrose2025elicitinglatentpredictionstransformers,nikankin2025taskdifferentcircuitsdisentangling}.
Output-level agreement therefore does not uniquely determine the internal computation learned by the student.
Here, textual prefixes serve as the states of the autoregressive decision process.
We use \emph{latent states} for the network's internal hidden representations.
Output supervision constrains predictions but only indirectly constrains the latent states used to produce them.
Particularly for complex reasoning tasks, output-level supervision may therefore leave valuable intermediate computational information untransferred~\citep{jiao-etal-2020-tinybert, sun-etal-2019-patient,wang2020minilmdeepselfattentiondistillation,wang2021minilmv2multiheadselfattentionrelation}. 

\begin{wrapfigure}{R}{0.48\textwidth}
      \centering
      \captionsetup{skip=4pt}
      \captionsetup[subfigure]{skip=1pt}

      \begin{subfigure}{\linewidth}
          \centering
          \includegraphics[width=\linewidth]
              {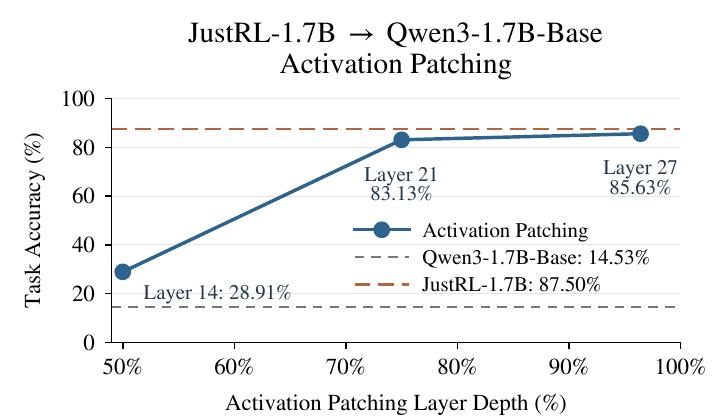}
          \caption{}
          \label{fig:hidden_replacement}
      \end{subfigure}

      \par\vspace{2pt}

      \begin{subfigure}{\linewidth}
          \centering
          \includegraphics[width=\linewidth]
              {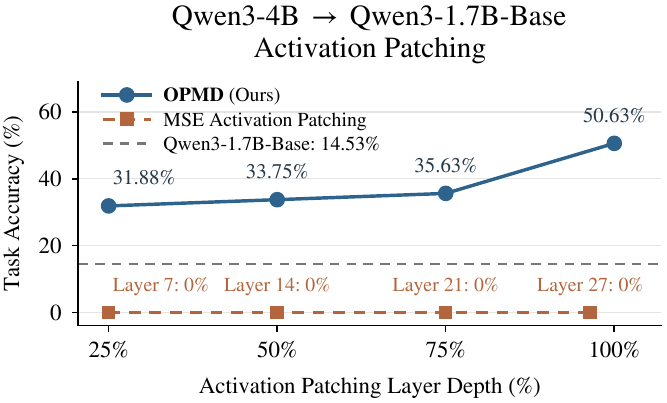}
          \caption{}
          \label{fig:projector_compatibility}
      \end{subfigure}

      \caption{
        Teacher latent states and student compatibility.
        AMC accuracy under interventions at different depths.
      }
      \label{fig:hidden_transfer_motivation}
  \end{wrapfigure}

This raises a natural question: \textit{can the latent states of a stronger teacher provide actionable information that improves the student's subsequent computation}?
Our preliminary analysis provides direct evidence that intermediate latent states contain actionable information for reasoning.
We evaluate on AMC 2023~\citep{numina_math_datasets} and use cross-model activation patching~\citep{prakash2024finetuning}, replacing the Base model's hidden state at a selected layer with the corresponding state from its RL-optimized counterpart while leaving all subsequent computation unchanged.
As shown in Figure~\ref{fig:hidden_transfer_motivation}\subref{fig:hidden_replacement}, patching latent states from an RL-optimized model into its base model substantially improves downstream task accuracy, demonstrating that RL-optimized latent states contain task-relevant information that remains usable by the Base model's downstream computation.

However, simply transferring such information is non-trivial, especially when teacher and student 
have different hidden dimensions or representation geometries, typically requiring a mapping between their representation spaces.
Existing feature-based distillation methods therefore rely on manually specified layer correspondences, learned projections, or specially designed relational objectives to make their intermediate latent states comparable \citep{jiao-etal-2020-tinybert, wang2021minilmv2multiheadselfattentionrelation, liang2023moretaskawarelayerwisedistillation}.
We further extend cross-model activation patching to heterogeneous teacher--student pairs by mapping the teacher's latent state into the student's hidden space.
As shown in Figure~\ref{fig:hidden_transfer_motivation}\subref{fig:projector_compatibility}, patching teacher latent states mapped by an MSE-trained projector can severely disrupt student generation, whereas states produced by a task-trained projector yield substantial gains across patching depths.
These results suggest that minimizing hidden-state reconstruction error is not sufficient to produce representations that the student's downstream computation can use.
What matters is therefore not only whether a teacher latent state can be mapped close to a student latent state, but whether the mapped latent state lies in a form that is computationally usable by the student's remaining layers.
This motivates constructing teacher-informed latent states directly within the student's computational interface.

Motivated by these observations, we do not directly force the student to reproduce the teacher's hidden states.
Instead, we first construct a \emph{student-compatible teacher} that transforms teacher information into intermediate states optimized to improve the student's downstream behavior and to be effectively used by its subsequent layers.
These states then serve as richer computational targets for distillation.
Concretely, a learned bridge injects teacher hidden states into a frozen student backbone, producing teacher-informed representations within the student's own computational interface.
We then perform \emph{on-policy internalization}, training a native student to reproduce these teacher-induced hidden trajectories on student-generated prefixes.
At deployment, the teacher model and bridge are removed, leaving only the native student.
This construction lets us test whether teacher-derived information becomes usable through the student's own parameters.

We evaluate \method{} on downstream reasoning tasks across three teacher--student pairs, demonstrating gains over output-level OPD.
We further extend the framework to multiple domain teachers and validate mechanism distillation through cross-family SFT without logit-level KL supervision.
Layer-level interventions further reveal changes in how students approach a problem, beyond changes in answer accuracy.
In a case study, intermediate-layer updates recover solution-route preferences closer to the fully distilled model, while a substantial accuracy gap remains.
These observations distinguish learning a preferred solution strategy from realizing the computation needed to execute it successfully.

Our contributions are threefold.
First, we construct a student-compatible teacher and train the native student to internalize its latent states through on-policy supervision.
Second, we evaluate this framework across teacher--student pairs, multiple domain teachers, and cross-family SFT.
Third, we investigate whether different layer groups internalize different reasoning mechanisms, using their distinct solution-route preferences to examine how strategy selection relates to successful execution.

\section{Related Work}
\label{sec:related_work}

\paragraph{On-policy language-model distillation.}
To mitigate exposure bias and distribution shift in autoregressive generation, on-policy distillation supervises the student on prefixes sampled from its own rollout policy \citep{gu2024minillm,agarwal2024onpolicy}, with recent variants exploring alternative divergence formulations and token-level feedback allocation \citep{xu2026tip,liu2026extremelysparse}.
Most existing methods operate through output-level teacher signals, improving where and how the student receives supervision without directly constraining its intermediate computation.
Recent work has begun extending the on-policy paradigm to representation-level supervision~\citep{yang2026oprdonpolicyrepresentationdistillation}.
\method{} similarly adopts student-generated prefixes, but focuses on constructing teacher-informed intermediate targets that can be effectively used by the student's subsequent computation.

\paragraph{Representation-based distillation.}
Intermediate supervision ranges from hidden-state and layer-wise matching to attention and relational distillation~\citep{romero2015fitnets,sun-etal-2019-patient,jiao2020tinybert,wang2020minilmdeepselfattentiondistillation,wang2021minilmv2multiheadselfattentionrelation,liang2023moretaskawarelayerwisedistillation}.
In reasoning models, OPRD aligns intermediate representations on student-generated trajectories, whereas PHF transfers hidden transitions and trajectory geometry from a reference-conditioned privileged teacher~\citep{yang2026oprdonpolicyrepresentationdistillation,li2026phf}.
These approaches demonstrate the value of intermediate supervision, but primarily define transfer through representational correspondence or trajectory matching, without explicitly ensuring that the resulting states can be effectively used by the student's subsequent computation.
\method{} addresses this distinction by constructing student-compatible intermediate targets before internalizing them into the native student.

\paragraph{Latent Interventions and Internalization.}
Intermediate computation can be modified by externally supplied signals.
Activation patching replaces internal states to measure their causal effect on subsequent computation, while learned conditioning architectures such as ControlNet inject additional information into frozen pretrained backbones through trainable control pathways~\citep{heimersheim2024useinterpretactivationpatching,pmlr-v235-ghandeharioun24a,zhang2023addingconditionalcontroltexttoimage}.
A complementary line of work internalizes privileged or contextual information so that it is no longer required at inference time~\citep{zhao2026selfdistilledreasoneronpolicyselfdistillation,ye2026onpolicycontextdistillationlanguage}.
\method{} combines these perspectives by using teacher-conditioned intervention to construct useful intermediate targets and subsequently internalizing the resulting hidden trajectories into the native student.

\section{Method}
\label{sec:method}

\method{} goes beyond direct representation matching by transferring teacher-derived intermediate information according to its effect on the student's subsequent computation.
This raises two distinct requirements: the transferred information must be expressed in a form that the student can productively use, and the resulting computation must ultimately be realizable by the student itself.
\method{} addresses these requirements through student-compatible teacher construction followed by on-policy internalization.

\begin{figure}[t]
    \centering
    \includegraphics[width=\linewidth]{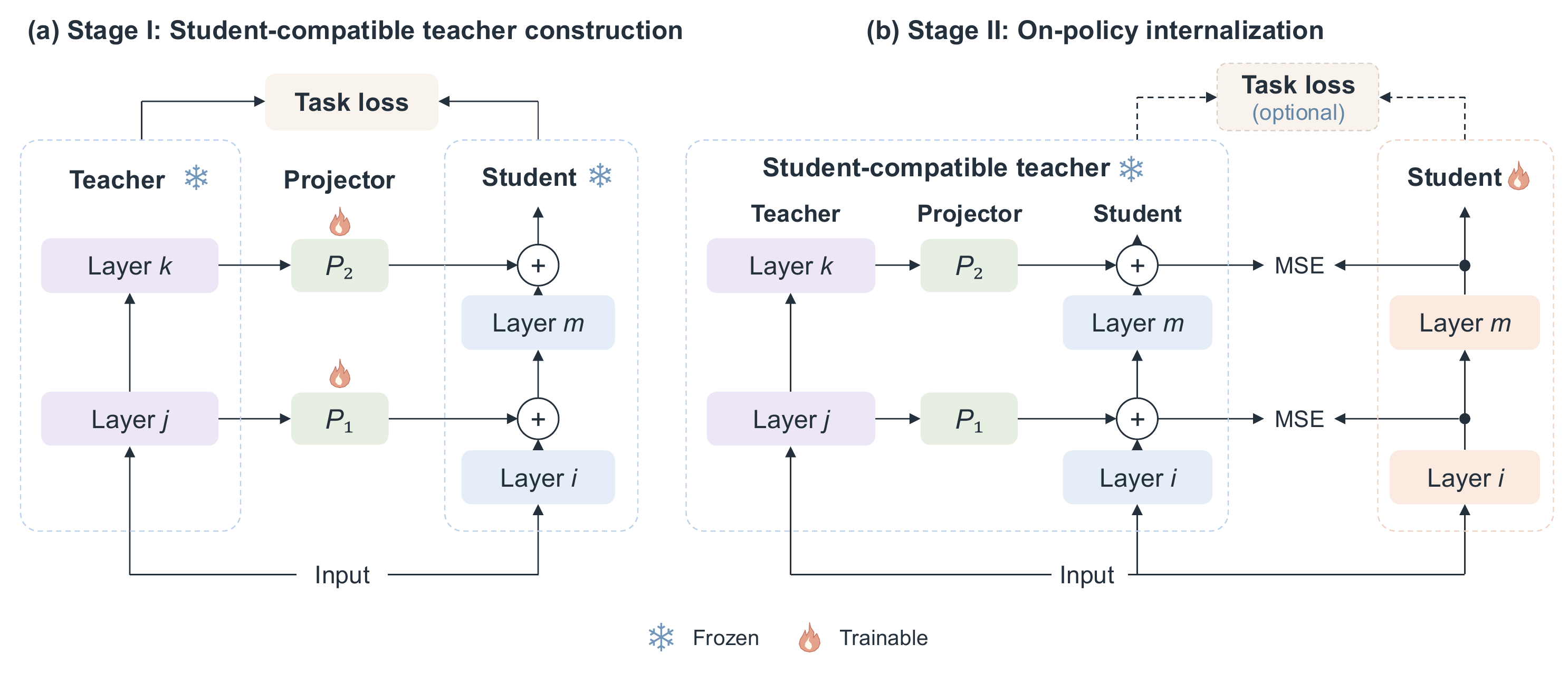}
    \caption{
        Overview of \method{}.
        Stage I learns layer-wise projectors to inject teacher representations into a frozen student, constructing a student-compatible teacher.
        Stage II performs on-policy internalization by aligning the native student's latent states with those of the student-compatible teacher.
    }
    \label{fig:pipeline}
\end{figure}

As illustrated in Fig.~\ref{fig:pipeline}, Stage I, student-compatible teacher construction, makes teacher information usable by the student's existing computation.
Stage II, on-policy internalization, trains the native student to reproduce the resulting latent states without external teacher intervention.
We address these problems sequentially.

\subsection{Preliminaries}

\paragraph{On-policy language-model distillation.}
Let $p_{\theta_T}$ and $p_{\theta_S}$ denote an autoregressive teacher model and a student model, parameterized by $\theta_T$ and $\theta_S$, respectively. 
On-policy distillation~\citep{gu2024minillm,agarwal2024onpolicy} aligns the student's predictions with the teacher's on student-generated prefixes.
Our OPD baseline uses reverse KL computed over the full vocabulary:

\begin{equation}
    \mathcal{L}_{\text{OPD}}(\theta_S)
    =
    \mathbb{E}_{x \sim \mathcal{D},\, \hat{y} \sim p_{\theta_S}(\cdot \mid x)}
    \left[
        \sum_{t=1}^{|\hat{y}|}
        D_{\mathrm{KL}}\left(
            p_{\theta_S}(\cdot \mid c_t)
            \,\Vert\,
            p_{\theta_T}(\cdot \mid c_t)
        \right)
    \right],
\label{eq:opd}
\end{equation}

where $c_t = (x, \hat{y}_{<t})$ is the prompt concatenated with the student's generated prefix up to step $t$. 
While this objective resolves exposure bias by matching the training and deployment prefix distributions, it provides only output-level supervision.

\paragraph{Geometric hidden-state matching.}
To supervise intermediate computations, prior works \citep{wang2021minilmv2multiheadselfattentionrelation, yang2026oprdonpolicyrepresentationdistillation} attempt to match the hidden representations of the teacher $h_j^T(c)$ and student $h_i^S(c)$ at selected layers. 
When architectures differ, this typically involves learning a projection function $P$ to minimize a geometric distance metric, such as Mean Squared Error (MSE):
\begin{equation}
    \mathcal{L}_{\text{MSE}}(P) = \mathbb{E}_{c} \left[ \left\| P(h_j^T(c)) - h_i^S(c) \right\|_2^2 \right].
\label{eq:mse_matching}
\end{equation}

However, minimizing geometric distance in representation space does not ensure that the projected teacher state $P(h_j^T(c))$ is functionally meaningful or computationally realizable by the student's subsequent downstream layers. 
\method{} explicitly departs from Equation~\ref{eq:mse_matching} by optimizing the injection of teacher states based entirely on their \emph{downstream functional utility}.

\subsection{On-Policy Mechanism Distillation}
On-Policy Mechanism Distillation (\method{}) goes beyond direct representation matching by transferring teacher-derived intermediate information according to its effect on the student's subsequent computation.

\paragraph{Student-compatible teacher construction.}
Let $p_{\theta_T}$ and $p_{\theta_S^0}$ denote the teacher and initial student, respectively.
Both models remain frozen throughout Stage I.
For a causal text prefix $c$, let $h_j^T(c)$ denote the teacher latent state at layer $j$, and let $h_i^{S,-}(c)$ denote the student latent state immediately before an intervention at layer $i$.
When teacher and student use different tokenizers, $c$ denotes the same textual prefix independently tokenized by each model.

We define a set of teacher--student transfer-layer pairs $\mathcal{I}=\{(j_r,i_r)\}_{r=1}^{R}$.
For each pair, a learned projector $P_{\phi_r}:\mathbb{R}^{d_T}\rightarrow\mathbb{R}^{d_S}$ maps the teacher representation to a residual update in the student's hidden space:
\begin{equation}
    \tilde{h}_{i_r}^{S}(c) = h_{i_r}^{S,-}(c) + P_{\phi_r}\!\left(h_{j_r}^{T}(c)\right),
    \qquad (j_r,i_r)\in\mathcal{I}.
    \label{eq:hidden_transfer}
\end{equation}
The resulting composition of the frozen teacher, learned projectors, and frozen student backbone defines the \emph{student-compatible teacher}, whose predictive distribution we denote by $\tilde{p}_{\phi}(\cdot\mid c)$.
Because the student backbone is fixed, the projectors must express teacher information in a form that can be acted on by the student's existing downstream computation.

\paragraph{Projector optimization.}
We optimize only the projector parameters $\phi$, while keeping both the teacher and student backbone frozen.
Rather than directly matching teacher and student hidden states, we train the projectors through the output behavior induced by the frozen student.
When teacher and student share a tokenizer, we minimize reverse KL over the full vocabulary
\begin{equation}
\mathcal{L}_{\mathrm{build}}(\phi)
=
\mathbb{E}_{c\sim\mathcal{C}}
\left[
D_{\mathrm{KL}}
\left(
\tilde{p}_{\phi}(\cdot\mid c)
\,\Vert\,
p_{\theta_T}(\cdot\mid c)
\right)
\right],
\label{eq:build}
\end{equation}
where $\mathcal{C}$ denotes the training contexts.
As in the OPD baseline, each KL term sums over the full vocabulary rather than using only the sampled token.
For different tokenizers, we can instead train the projectors with SFT on teacher-generated responses tokenized by the student, keeping both backbones frozen.
Appendix~\ref{sec:optimization_objectives} specifies these alternatives.

\paragraph{On-policy internalization.}
The student-compatible teacher still relies on the external teacher and projectors.
We train a native student $p_{\theta_S}$ initialized from $p_{\theta_S^0}$ to reproduce its intermediate states without either component.
Our on-policy experiments update student LoRA~\citep{hu2022lora} parameters while keeping the backbone frozen.
For each prompt $x\sim\mathcal{D}$, the current student samples $\hat{y}\sim p_{\theta_S}(\cdot\mid x)$.
The native student and the frozen student-compatible teacher are then evaluated on the same student-generated textual prefixes.

Let $c_t^{+}=(x,\hat{y}_{\le t})$ denote the prefix including response token $\hat{y}_t$.
We match the native student's latent state $h_i^S(c_t^{+};\theta_S)$ and the student-compatible teacher's latent state $\tilde{h}_i^S(c_t^{+})$ at each response-token position without shifting.
\begin{equation}
\mathcal{L}_{\mathrm{int}}(\theta_S)
=
\mathbb{E}_{
x\sim\mathcal{D},
\,\hat{y}\sim p_{\theta_S}(\cdot\mid x)
}
\left[
\frac{1}{|\mathcal{H}|\,|\hat{y}|}
\sum_{i\in\mathcal{H}}
\sum_{t=1}^{|\hat{y}|}
\frac{1}{d_S}
\left\|
h_i^S(c_t^{+};\theta_S)
-
\operatorname{sg}
\left(
\tilde{h}_i^S(c_t^{+})
\right)
\right\|_2^2
\right],
\label{eq:internalization}
\end{equation}
where $\operatorname{sg}(\cdot)$ denotes stop-gradient.
Because the targets are produced within the frozen student backbone, they already lie in the student's representation space and require no additional cross-model mapping.
Internalization supports a joint downstream task loss, but our on-policy experiments use only the latent-state objective in this stage.
SFT instead combines mechanism supervision on teacher-response prefixes with supervised token prediction, without logit-level KL supervision.
At inference time, both the teacher and projectors are discarded, leaving only the native student.

\section{Experiments}
\label{sec:experiments}

We evaluate task performance, transfer across training settings, and the behavioral changes retained by weight patching.
To facilitate reproducibility, we provide training hyperparameters, implementation details, and computational resources in Appendix~\ref{sec:training_details}.
We will release the code upon acceptance.

\subsection{Does On-Policy Internalization Improve Distillation?}
We first test whether \method{} improves mathematical reasoning over output-level distillation across different teachers and students.

\paragraph{Setting and baselines.}
Our main mathematical distillation experiments use DAPO-Math-17K~\citep{yu2025dapo} for training and AMC, AIME 2024~\citep{numina_math_datasets}, and AIME 2025~\citep{balunovic2026matharena} for evaluation.
The teachers are Qwen3-4B~\citep{yang2025qwen3technicalreport} and Qwen3-4B-Non-Thinking-RL-Math (RL-Math)~\citep{yang2026learningteachergeneralizedonpolicy}.
We use Qwen3-1.7B-Base (abbreviated as Qwen3-1.7B) and Qwen2.5-3B-Instruct (abbreviated as Qwen2.5-3B)~\citep{qwen2025qwen25technicalreport} as students in the main comparisons.
The main comparisons are Qwen3-4B $\rightarrow$ Qwen3-1.7B-Base, RL-Math $\rightarrow$ Qwen3-1.7B-Base, and RL-Math $\rightarrow$ Qwen2.5-3B-Instruct.
Within each comparison, methods share the student initialization and training data and are evaluated at matched training steps.
We compare against output-level OPD~\citep{gu2024minillm,agarwal2024onpolicy,li2026rethinkingonpolicydistillationlarge}, SFT on teacher-generated responses, and OPRD-Bridge~\citep{yang2026oprdonpolicyrepresentationdistillation}, which uses a frozen low-rank alignment for representation distillation across models.

\paragraph{Evaluation protocol.}
The main mathematical comparisons report avg@16, the mean final-answer accuracy over 16 sampled responses per problem.
The checkpoint analyses evaluate 40 AMC problems and 30 problems per AIME benchmark with temperature $0.7$ and top-$p=0.95$.
Appendix~\ref{sec:checkpoint_protocol} specifies experiment-specific evaluation settings.

\paragraph{Training protocol.}
In Stage I, \method{} trains its projectors at $5\times10^{-6}$ with both backbones frozen, then freezes the constructed teacher for Stage II.
The comparisons update student LoRA parameters with rank and scaling factor 64, zero dropout, and AdamW at learning rate $10^{-6}$.
In the Qwen3-4B to Qwen3-1.7B-Base run, each rollout collection contains 64 prompts with four responses each, sampled at temperature $1.0$ and top-$p=1.0$.
The prompt, response, and total training sequence limits are 1,024, 7,168, and 8,192 tokens, respectively.
Each optimizer update uses eight responses, giving 32 updates per rollout collection.
Stage II of \method{} uses latent-state supervision without a downstream task loss, whereas stage I of \method{} and OPD use full-vocabulary reverse KL.
Tables~\ref{tab:core_training_hyperparameters} and~\ref{tab:training_steps_by_setting} report the shared hyperparameters and setting-specific update counts.

\paragraph{Main results.}
Table~\ref{tab:main_results} shows AMC gains over OPD of 3.76, 7.04, and 2.66 percentage points, respectively.
With RL-Math and Qwen3-1.7B-Base, \method{} also improves AIME 2024 and AIME 2025 by 4.37 and 3.95 points.
Changing the teacher from Qwen3-4B to RL-Math lowers OPD's AMC accuracy from 35.31\% to 33.12\%, but raises \method{}'s from 39.07\% to 40.16\%.
Thus, a stronger teacher does not automatically yield a stronger distilled student, and \method{} makes better use of the specialized teacher in this comparison.

\begin{table}[t]
    \centering
    \caption{
        Mathematical reasoning accuracy (\%) across three teacher--student settings.
        All distillation methods use DAPO-Math-17K and are evaluated after 512 student updates, corresponding to Stage II for \method{}.
        Bold indicates the best reported student result in each column.
    }
    \label{tab:main_results}

    \small
    \setlength{\tabcolsep}{3.2pt}
    \renewcommand{\arraystretch}{1.12}

    \begin{tabular}{
        @{}l
        ccc
        @{\hspace{7pt}}
        ccc
        @{\hspace{7pt}}
        ccc
        @{}
    }
        \toprule
        & \multicolumn{3}{c}{\textbf{Qwen3-4B} $\rightarrow$ \textbf{Qwen3-1.7B}}
        & \multicolumn{3}{c}{\textbf{RL-Math} $\rightarrow$ \textbf{Qwen3-1.7B}}
        & \multicolumn{3}{c}{\textbf{RL-Math} $\rightarrow$ \textbf{Qwen2.5-3B}} \\
        \cmidrule(lr){2-4}
        \cmidrule(lr){5-7}
        \cmidrule(l){8-10}

        \textbf{Method}
        & AMC & AIME '24 & AIME '25
        & AMC & AIME '24 & AIME '25
        & AMC & AIME '24 & AIME '25 \\
        \midrule

        \multicolumn{10}{@{}l}{\textit{Reference models}} \\[-2pt]
        Teacher
        & 66.09 & 25.62 & 17.50
        & 91.56 & 56.25 & 46.67
        & 91.56 & 56.25 & 46.67 \\
        Student
        & 14.53 & 1.25 & 2.08
        & 14.53 & 1.25 & 2.08
        & 37.19 & 6.88 & 2.08 \\

        \addlinespace[2pt]
        \midrule

        \multicolumn{10}{@{}l}{\textit{Distillation methods}} \\[-2pt]

        SFT
        & 27.34 & 3.12 & 3.75
        & 29.06 & 3.54 & 3.75
        & 38.28 & 5.21 & \textbf{3.70} \\

        OPD
        & 35.31 & \textbf{8.33} & 3.96
        & 33.12 & 5.42 & 3.13
        & 38.75 & 6.88 & 3.12 \\

        OPRD-Bridge
        & 33.28 & 5.83 & 2.92
        & 29.68 & 7.92 & 2.92
        & 38.91 & \textbf{7.92} & 2.71 \\

        \method{}
        & \textbf{39.07} & \textbf{8.33} & \textbf{5.20}
        & \textbf{40.16} & \textbf{9.79} & \textbf{7.08}
        & \textbf{41.41} & 7.08 & 2.92 \\

        \bottomrule
    \end{tabular}
\end{table}

\begin{figure}[H]
    \centering
    \includegraphics[width=0.94\linewidth]{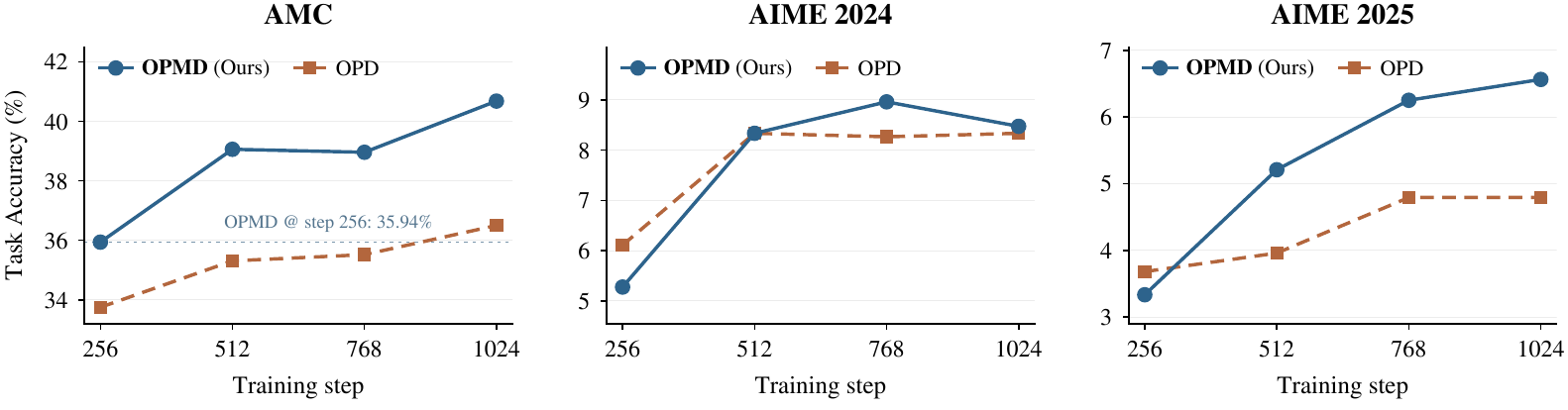}
    \caption{Accuracy through 1,024 student updates on AMC, AIME 2024, and AIME 2025.
    We compare \method{} and OPD using means over available seeds.}
    \label{fig:training_step_accuracy}
\end{figure}

\paragraph{Further gains from continued internalization.}
We next examine whether the 512-update in Table~\ref{tab:main_results} comparison captures the full benefit of internalization by extending student training to 1,024 updates.
Figure~\ref{fig:training_step_accuracy} compares \method{} and OPD across this longer trajectory, with updates counted within Stage II for \method{}.
The curves summarize available-seed means from the trajectory runs.

On AMC and AIME 2025, \method{} reaches its highest observed accuracy at 1,024 updates, exceeding its performance at the 512-update comparison point.
The continued gains show that the main table does not exhaust the benefit of internalization and that additional student training can further improve performance.
AIME 2024 instead peaks at 768 updates, indicating that the benefit of longer training depends on the task.

\noindent\begin{minipage}{\linewidth}
\begin{minipage}[t]{\dimexpr0.55\linewidth-\columnsep\relax}
\vspace{0pt}
\paragraph{Accuracy beyond longer responses.}
We also examine whether these gains simply reflect generating more tokens.
Figure~\ref{fig:mechanism_analysis_response_length} shows that OPD responses become substantially longer during training, while its AMC accuracy improves relatively modestly.
\method{} also generates longer responses as training proceeds, but achieves higher accuracy with shorter responses at every evaluated update count.
Across the overlapping response-length range, its trajectory remains above OPD's, indicating better accuracy at comparable token counts.
Even at 256 updates, \method{} exceeds OPD's accuracy at 512 and 768 updates while generating fewer tokens.
Both methods use the same 7,168-token evaluation limit, so this difference is not imposed by a smaller generation budget for \method{}.
The resulting advantage concerns both answer quality and output-token cost, which accuracy alone would not capture.
Together, the two figures show that continued internalization improves the effectiveness of the generated reasoning, beyond merely increasing response length.
Appendix~\ref{sec:checkpoint_protocol} provides the evaluation settings.
\end{minipage}\hfill%
\begin{minipage}[t]{0.45\linewidth}
\vspace{0pt}
    \centering
    \includegraphics[width=\linewidth]{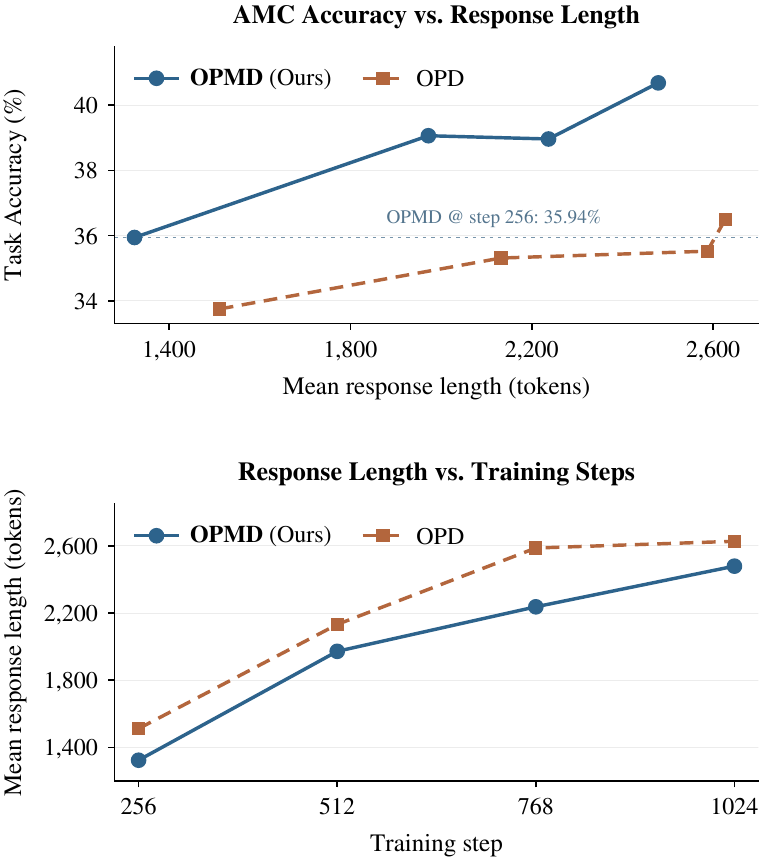}
    \captionof{figure}{AMC accuracy versus mean response length (top) and response length across training checkpoints (bottom).}
    \label{fig:mechanism_analysis_response_length}
\end{minipage}
\end{minipage}
\par

We further test whether direct representation matching can replace student-compatible teacher construction (Appendix~\ref{sec:projected_mse_ablation}).
Even with projector prefitting, OPD with projected MSE loses accuracy during continued training, suggesting that initial alignment alone does not ensure stable transfer, whereas Stage II of \method{} uses latent-state supervision without an OPD loss.

\subsection{Extensibility to Multi-Teacher and SFT Regimes}
\label{sec:multi_sft}

\begin{table}[!hb]
\centering
\begin{minipage}[t]{0.49\linewidth}
    \centering
    \footnotesize
    \setlength{\tabcolsep}{3pt}
    \renewcommand{\arraystretch}{1.12}
    \caption{
        Two-teacher distillation.
        Teacher scores use RL-Math on AMC and RL-Code on code tasks.
        Bold denotes the best student result.
    }
    \label{tab:two_teacher_results}

    \begin{tabular*}{\linewidth}{@{\extracolsep{\fill}}lccc@{}}
        \toprule
        & \multicolumn{3}{c}{\textbf{Student: Qwen3-1.7B}} \\
        \cmidrule(l){2-4}
        \textbf{Model / method} & AMC & HumanEval+ & MBPP+ \\
        \midrule
        \multicolumn{4}{@{}l}{\textit{Reference models}} \\
        \quad Domain Teacher & 91.56 & 86.00 & 70.20 \\
        \quad Student & 14.53 & 1.83 & 6.61 \\
        \midrule
        \multicolumn{4}{@{}l}{\textit{Distilled students}} \\
        \quad M-OPD       & 33.91 & 59.76 & 50.53 \\
        \quad \method{}   & \textbf{38.22} & \textbf{60.37}
                          & \textbf{57.14} \\
        \bottomrule
    \end{tabular*}
\end{minipage}\hfill
\begin{minipage}[t]{0.49\linewidth}
    \centering
    \footnotesize
    \setlength{\tabcolsep}{4pt}
    \renewcommand{\arraystretch}{1.12}

    \caption{
        Cross-family SFT with mechanism distillation.
        Scores are accuracies (\%).
        Bold denotes the best non-teacher result.
    }
    \label{tab:textual_sft_distillation}

    \begin{tabular*}{\linewidth}{@{\extracolsep{\fill}}lccc@{}}
        \toprule
        & \multicolumn{3}{c}{\textbf{Qwen3-4B} $\rightarrow$ \textbf{Llama3.1-8B}} \\
        \cmidrule(l){2-4}
        \textbf{Method} & AMC & AIME'24 & AIME'25 \\
        \midrule

        \textit{Reference models} \\
        \quad Teacher & 66.09 & 25.62 & 17.50 \\
        \quad Student & 19.64 & \textbf{4.72} & 0.63 \\
        
        \midrule

        \textit{Textual distillation} \\
        \quad SFT & 22.50 & 4.33 & 1.46 \\
        \quad SFT+MD & \textbf{25.31} & 3.88 & \textbf{1.81} \\ 
        
        \bottomrule
    \end{tabular*}
\end{minipage}
\end{table}

\paragraph{Distillation from two domain teachers.}
We test whether one student can benefit from teachers specialized in different domains.
We use Qwen3-4B-Non-Thinking-RL-Code (RL-Code) alongside RL-Math~\citep{yang2026learningteachergeneralizedonpolicy}.
We distill RL-Math-Step500 and RL-Code-Step300 into Qwen3-1.7B-Base using G-OPD-Training-Data~\citep{yang2026learningteachergeneralizedonpolicy}, with 25,276 prompts per domain.
Each prompt is routed to its domain teacher and corresponding projector bank, without enforcing an equal domain split within every minibatch.
Stage I and Stage II use 128 and 512 updates, respectively.
We evaluate on HumanEval+~\citep{chen2021evaluatinglargelanguagemodels,liu2023is} and MBPP+~\citep{austin2021programsynthesislargelanguage,liu2023is}, using 164 and 378 problems, respectively, with the augmented tests provided by EvalPlus.
Evaluation uses one greedy response per problem and reports pass@1.
Relative to M-OPD, \method{} improves AMC, HumanEval+, and MBPP+ by 4.31, 0.61, and 6.61 percentage points (Table~\ref{tab:two_teacher_results}).
The shared student improves on both domains, so the measured math gain does not come at the expense of code accuracy relative to M-OPD.
The larger gain on MBPP+ than on HumanEval+ also shows that the benefit varies across tasks, rather than providing a uniform improvement from adding mechanism supervision.

\paragraph{Validating mechanism distillation through SFT.}
We test whether mechanism distillation improves student performance without logit-level KL supervision.
Using Qwen3-4B $\rightarrow$ Llama3.1-8B-Instruct~\citep{grattafiori2024llama3herdmodels}, we compare SFT and SFT+MD on the same teacher-response replay data.
Both methods train for 256 updates with batch size 16, learning rate $5\times10^{-6}$, and an 8,192-token sequence limit.
SFT+MD adds latent-state MSE with coefficient $0.5$ to the response NLL, using frozen targets from the 128-update student-compatible teacher.
Each model uses its native chat template, with teacher states aligned to the response prefix visible to the student.
Neither condition uses a KL loss on teacher and student logits.
Table~\ref{tab:textual_sft_distillation} shows improvements on AMC (25.31\% vs.\ 22.50\%) and AIME 2025 (1.81\% vs.\ 1.46\%), but a decrease on AIME 2024 (3.88\% vs.\ 4.33\%).
The gains show that mechanism supervision can add useful information beyond the teacher's textual responses, without relying on a logit-level distillation objective.

\subsection{What Do Different Layers Learn?}
\label{sec:different_layers}
We ask whether different layer groups internalize different reasoning mechanisms.
Here, OPMD student denotes the distilled Qwen3-1.7B-Base student at step 1024, retaining all trained parameters and requiring no teacher or projectors at inference.
Cross-model activation patching substitutes activations between original and fine-tuned models~\citep{prakash2024finetuning}.
Weight patching instead copies selected distilled layer parameters into the raw student, keeping other parameters fixed~\citep{pmlr-v267-oozeer25a}.
Patched models compute their own activations.

\begin{table}[H]
\centering
\begin{minipage}[t]{0.49\linewidth}
\centering\footnotesize
\setlength{\tabcolsep}{4pt}
\caption{
    Layer-wise weight patching into the Qwen3-1.7B-Base student.
    Each slice uses the corresponding layers from \method{} at step 1024.
    Bold marks the highest accuracy and lowest token count among patched models.
}
\label{tab:opmd_layer_splice}
\begin{tabular*}{\linewidth}{@{\extracolsep{\fill}}lrr@{}}
    \toprule
    \multicolumn{3}{c}{Weight patching in Qwen3-1.7B-Base} \\
    \shortstack[l]{Model\\variant}
    & Acc (\%) $\uparrow$
    & Toks $\downarrow$ \\
    \midrule
    \multicolumn{3}{@{}l}{\textit{Reference models}} \\
    Raw student
    & 14.84 & 3147 \\
    OPMD student & 40.68 & 2462 \\
    \midrule
    \multicolumn{3}{@{}l}{\textit{Patched models}} \\
    Patch 1--7
    & 31.88 & \textbf{1785} \\
    Patch 8--14
    & \textbf{34.06} & 1936 \\
    Patch 15--28$^{\dagger}$
    & 23.59 & 2494 \\
    \bottomrule
\end{tabular*}
\end{minipage}\hfill
\begin{minipage}[t]{0.49\linewidth}
\centering\footnotesize
\setlength{\tabcolsep}{2pt}
\caption{Solution routes on two selected AMC problems.
Counts are out of all 16 responses, including residual cases.
Accuracy follows the archived final-answer grader.
Bold marks the highest answer accuracy for each problem.}
\label{tab:amc-splice-routes}
\begin{tabular*}{\linewidth}{@{\extracolsep{\fill}}lrrr@{}}
\toprule
\shortstack[l]{Solution strategy\\/ accuracy} & Patch 1--7 & Patch 8--14 & \shortstack{OPMD\\student} \\
\midrule
\multicolumn{4}{@{}l}{\textit{q17}} \\
Direct route & 12/16 & 6/16 & 5/16 \\
Mediated route & 4/16 & 9/16 & 11/16 \\
Answer accuracy & 18.75\% & 62.50\% & \textbf{81.25\%} \\
\midrule
\multicolumn{4}{@{}l}{\textit{q22}} \\
Direct route & 7/16 & 5/16 & 5/16 \\
Mediated route & 5/16 & 10/16 & 11/16 \\
Answer accuracy & 18.75\% & 31.25\% & \textbf{50.00\%} \\
\bottomrule
\end{tabular*}
\end{minipage}
\end{table}

\paragraph{Recovering accuracy through weight patching.}
We test whether selected distilled blocks retain the OPMD student's accuracy gains.
We patch the weights of blocks 1 to 7 (Patch 1--7) or 8 to 14 (Patch 8--14) from the step-1024 \method{} checkpoint into raw Qwen3-1.7B-Base, leaving the remaining parameters unchanged.
Table~\ref{tab:opmd_layer_splice} shows that Patch 1--7 and Patch 8--14 improve accuracy from 14.84\% to 31.88\% and 34.06\%, while reducing mean response length.
Both produce shorter responses than Full but remain below its 40.68\% accuracy, separating brevity from accuracy.

\paragraph{Case study: solution routes after weight patching.}
We test whether patched models recover Full's solution strategies despite lower accuracy.
We first selected q17 and q22 for Full's archived accuracy advantage over Q1 of 62.50 and 31.25 points, respectively.
GPT 5.6 luna then independently relabeled all 96 responses with identities, prior labels, correctness, and expected conclusions withheld.
All 16 responses per model and problem were retained.

Direct routes solve from the problem's structure or transfer constraints, whereas mediated routes use a logarithmic calculation on q17 or a separately derived common liquid level on q22.
Labels describe the adopted solution procedure, regardless of answer correctness.
Q1 favors direct routes, while Q2 and Full favor mediated routes (Table~\ref{tab:amc-splice-routes}).
Q2's closer route preferences but lower accuracy suggest that different layer groups retain different aspects of the reasoning mechanism, with strategy selection only partly recovering successful execution.
These cases do not establish exclusive functions for either layer group.
Appendix~\ref{sec:route_case_protocol} provides the annotation protocol and definitions, with response comparisons in Appendix~\ref{sec:route_examples}.

\subsection{Accuracy vs. Training Cost}
\label{sec:training_efficiency}
\noindent\begin{minipage}{\linewidth}
\begin{minipage}[t]{\dimexpr0.50\linewidth-\columnsep\relax}
\vspace{0pt}
We test whether \method{} improves the accuracy--cost trade-off after accounting for teacher construction.
Using Qwen3-4B $\rightarrow$ Qwen3-1.7B-Base on two NVIDIA H100 GPUs, we compare AMC accuracy against training time and FLOPs (Figure~\ref{fig:accuracy_training_cost}).
OPMD reaches 38.96\% accuracy at 768 student updates, compared with 36.50\% for OPD at 1,024 updates.
The plotted OPMD costs include Stage I, so the comparison accounts for teacher construction as well as student training.
With our optimized vLLM implementation~\citep{kwon2023efficientmemorymanagementlarge}, Stage II of \method{} achieves nearly the same update throughput as OPD.
We also examine shorter Stage I runs in Appendix~\ref{sec:stage1_sensitivity}.
Shorter teacher construction in Stage I can retain comparable downstream accuracy, with RL-Math, 32 Stage I updates yield 40.00\% AMC accuracy, compared with 40.16\% after 96 updates.
\end{minipage}\hfill%
\begin{minipage}[t]{0.5\linewidth}
\vspace{0pt}
\centering
\includegraphics[width=\linewidth]{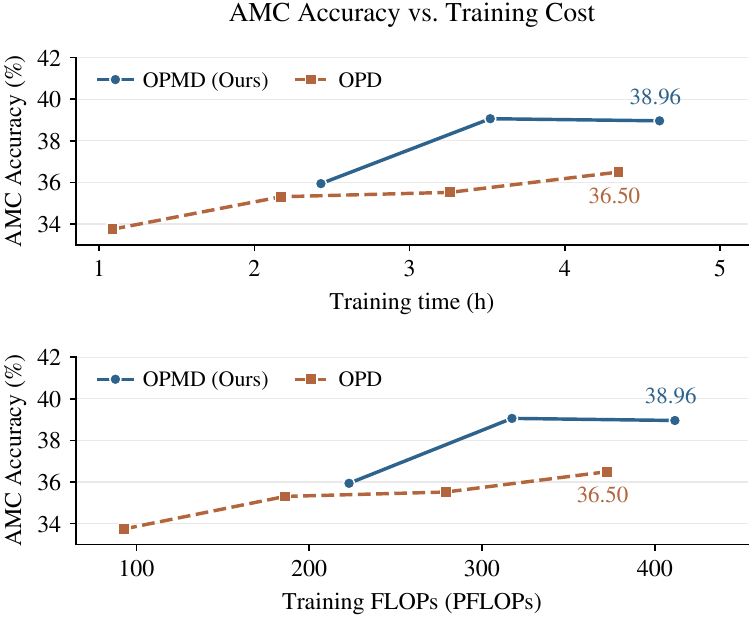}
\captionof{figure}{AMC accuracy versus training time (top) and FLOPs (bottom).
OPMD costs include Stage I and Stage II.}
\label{fig:accuracy_training_cost}
\end{minipage}
\end{minipage}
\par

\section{Conclusion}
We introduced On-policy Mechanism Distillation (OPMD), an on-policy distillation framework that moves beyond output supervision by explicitly transferring intermediate computation mechanisms.
Our key observation is that directly matching teacher and student representations is not sufficient, geometric alignment does not guarantee that transferred states can be effectively used by the student's subsequent computation.
OPMD addresses this gap in two stages, it first constructs a student-compatible teacher by optimizing teacher-derived intermediate states for their downstream functional utility, and then internalizes these states on student-generated trajectories so that the resulting student operates independently at inference time.
Across three teacher-student settings, OPMD significantly improves reasoning capability over output-level OPD, while also extending to multi-teacher distillation and cross-family supervised fine-tuning.
Further analyses show that these gains are accompanied by more effective reasoning at comparable or lower response cost, and that the internalized updates induce meaningful changes in intermediate computation and solution strategies rather than merely reproducing final outputs.
Taken together, our results establish mechanism-level supervision as a practical extension of on-policy distillation beyond output matching, and highlight computational compatibility and internalization as two key principles for transferring how a stronger model reasons, not only what it predicts.

\section*{Limitations}
Our evaluation covers a limited set of teacher--student pairs and downstream tasks.
Whether the observed gains extend to larger models, a broader range of architectures, and tasks beyond mathematics and code generation remains to be tested.
The two cases in Section~\ref{sec:different_layers} were selected for the OPMD student's advantage over Patch 1--7 and do not establish general layer specialization or causal mechanisms.

\section*{AI Use Statement}
Generative AI tools assisted with manuscript editing, and experimental reporting.
GPT 5.6 luna was also used to annotate solution routes in 96 model-generated responses for the case study, with model identities, correctness labels, prior annotations, and expected conclusions withheld.
Appendix~\ref{sec:route_case_protocol} documents the annotation procedure, disagreements, and exact prompt.
These annotations characterize solution routes and do not replace final-answer grading.
The authors take responsibility for the final text, methodology, claims, and reported results.

\section*{Reproducibility Statement}
Section~\ref{sec:method} specifies the construction and internalization objectives.
Appendix~\ref{sec:training_details} provides training hyperparameters, update counts, and experiment-specific implementation details.
Appendix~\ref{sec:checkpoint_protocol} describes the evaluation settings and case-selection procedure and includes the prompt used for route annotation.
Appendix~\ref{sec:additional_accuracy} defines the per-problem accuracy-change analysis and its filtering rule.

\section*{Ethics Statement}
This work studies language-model distillation using existing mathematical reasoning and code benchmarks, with behavioral analysis conducted on model-generated responses.
Distillation may transfer errors and biases from teachers, and improvements on these benchmarks do not establish reliability or safety in deployment.
The route annotations are model-generated judgments and are interpreted as descriptive evidence, with disagreements and limitations reported in Appendix~\ref{sec:route_case_protocol}.

\bibliography{iclr2027_conference}
\bibliographystyle{iclr2027_conference}

\newpage
\appendix
\section{Training and Implementation Details}
\label{sec:training_details}
Tables~\ref{tab:core_training_hyperparameters} and~\ref{tab:training_steps_by_setting} report the shared training hyperparameters and training steps for the four on-policy settings.
The main comparisons use 512 student-training steps, while the trajectory and layer analyses also examine later steps, including the step-1024 \method{} model.
\subsection{Optimization Objectives}
\label{sec:optimization_objectives}
\paragraph{Student-compatible teacher construction.}
For shared tokenizers, projector training uses the reverse KL objective in Equation~\ref{eq:build}.
For different tokenizers, let $\mathcal{D}_T$ contain prompts $x$ and teacher-generated response texts $y$, with response tokens $y_t^S$ defined by the student's tokenizer.
An alternative projector objective is
\begin{equation}
\mathcal{L}_{\mathrm{build}}^{\mathrm{SFT}}(\phi)
= -\mathbb{E}_{(x,y)\sim\mathcal{D}_T}
\left[\frac{1}{|y^S|}\sum_{t=1}^{|y^S|}
\log\tilde{p}_{\phi}(y_t^S\mid x,y_{<t}^S)\right].
\end{equation}
Only the projectors are updated, while both model backbones remain frozen.
The teacher supplies latent states for the corresponding textual context, without requiring a common output vocabulary.

\paragraph{Native student training.}
Our on-policy experiments use only $\mathcal{L}_{\mathrm{int}}$ on student-generated prefixes, without a downstream task loss.
For SFT+MD, we use the fixed weighted sum
\begin{equation}
\mathcal{L}_{\mathrm{SFT+MD}}=0.5\mathcal{L}_{\mathrm{rep}}+\mathcal{L}_{\mathrm{SFT}}.
\end{equation}
Here, $\mathcal{L}_{\mathrm{rep}}$ is the latent-state discrepancy evaluated on teacher-response prefixes, and $\mathcal{L}_{\mathrm{SFT}}$ is the supervised negative log-likelihood of those responses under the native student.
The representation and SFT coefficients are fixed to $0.5$ and $1$, respectively.
The SFT baseline uses only $\mathcal{L}_{\mathrm{SFT}}$, and neither SFT condition uses logit-level KL supervision.
The absence of a downstream task loss in on-policy native student training does not remove the output objective used for teacher construction.

\subsection{Hyperparameters}
\begin{table}[htbp]
\centering
\small
\caption{Core training hyperparameters shared across OPMD and OPD settings.}
\label{tab:core_training_hyperparameters}
\begin{tabular}{@{}lccc@{}}
\toprule
Setting & OPMD Stage I & OPMD Stage II & OPD \\
\midrule
Trainable parameters & Projectors & Student LoRA & Student LoRA \\
LoRA rank / alpha & -- & 64 / 64 & 64 / 64 \\
LoRA dropout & -- & 0 & 0 \\
Optimizer & AdamW & AdamW & AdamW \\
Learning rate & $5\times10^{-6}$ & $10^{-6}$ & $10^{-6}$ \\
LR schedule / warm-up & Constant / 0 & Constant / 0 & Constant / 0 \\
Batch size & 8 & 8 & 8 \\
Responses per prompt & 4 & 4 & 4 \\
\bottomrule
\end{tabular}
\end{table}

\begin{table}[htbp]
\centering
\small
\caption{Training steps across the four on-policy distillation settings.}
\label{tab:training_steps_by_setting}
\begin{tabular}{@{}lccc@{}}
\toprule
Teacher $\rightarrow$ Student & OPMD Stage I & OPMD Stage II & OPD \\
\midrule
Qwen3-4B $\rightarrow$ Qwen3-1.7B & 128 & 512 & 512 \\
RL-Math $\rightarrow$ Qwen3-1.7B & 96 & 512 & 512 \\
RL-Math $\rightarrow$ Qwen2.5-3B & 128 & 512 & 512 \\
RL-Math + RL-Code $\rightarrow$ Qwen3-1.7B & 128 & 512 & 512 \\
\bottomrule
\end{tabular}
\end{table}

\paragraph{Additional settings for multi-teacher and SFT experiments.}
Multi-teacher training uses the same 1,024-token prompt, 7,168-token response, and 8,192-token sequence limits as the Qwen3-4B main run.
Its Stage II MSE supervises all response positions at all 28 block outputs before the final RMSNorm.
The cross-family SFT replay pool retains 8,188 of 8,192 records after excluding four truncated source prompts.
Both SFT conditions use LoRA rank and scaling factor 64, zero dropout, and AdamW with zero weight decay.
Response alignment uses only teacher tokens reproducible from the currently visible response prefix.

\section{Evaluation Details}
\label{sec:checkpoint_protocol}
Table~\ref{tab:evaluation_settings} records the confirmed evaluation settings and their scope.
The checkpoint curves average available repeated-generation evaluations for fixed checkpoints, rather than independent training runs.
Mathematical evaluations use 16 responses per problem at temperature $0.7$ and top-$p=0.95$.
Code evaluation uses temperature $0$, top-$p=1$, no top-$k$ restriction, and a maximum of 16,384 generated tokens.
EvalPlus~\citep{liu2023is} grades the extracted programs against its augmented test suites.
Main AIME evaluations use a 31,000-token generation limit.
AMC evaluations use 7,168 tokens with the Qwen3-4B teacher and 31,000 tokens in the other settings.
\begin{table}[htbp]
\centering
\small
\caption{Confirmed evaluation settings by experiment.}
\label{tab:evaluation_settings}
\begin{tabular}{@{}p{0.25\linewidth}p{0.35\linewidth}p{0.32\linewidth}@{}}
\toprule
Scope & Setting & Value \\
\midrule
AMC, Qwen3-4B teacher & Maximum response length & 7,168 tokens \\
AMC, other teacher settings & Maximum response length & 31,000 tokens \\
AIME, main settings & Maximum response length & 31,000 tokens \\
\midrule
Route case study & Distilled checkpoint & Step 1024 \\
 & Patched blocks & 1--7 or 8--14 \\
 & Responses per model / problem & 16 \\
 & Generation seed & 0 \\
 & Temperature / top-$p$ & 0.7 / 0.95 \\
 & Maximum response length & 31,000 tokens \\
 & Answer grading & Archived final-answer labels \\
\midrule
Independent route annotation & Annotator & GPT 5.6 luna \\
 & Reasoning effort & Medium \\
 & Responses / batches & 96 / 16 \\
\bottomrule
\end{tabular}
\end{table}

\subsection{Solution-Route Case Study}
\label{sec:route_case_protocol}

\paragraph{Selection and responses.}
We first compared archived answer accuracy across problems and selected q17 and q22 because the OPMD student clearly outperformed Patch 1--7.
The respective accuracies of the OPMD student and Patch 1--7 were 81.25\% and 18.75\% on q17, and 50.00\% and 18.75\% on q22.
Only after selecting these cases did we inspect the responses and compare their solution routes.
Patch 1--7 and Patch 8--14 patch the weights of blocks 1 to 7 and 8 to 14 from the step-1024 \method{} checkpoint into raw Qwen3-1.7B-Base.
All other parameters remain those of the raw student.
We retain all 16 archived responses per model and problem, including incorrect and incomplete outputs.
Accuracy follows the original final-answer grader and does not certify the correctness of each reasoning step.

\paragraph{Interpreting the routes.}
The identifiers q17 and q22 refer to archived examples rather than official AMC topic labels.
The shared terms \emph{direct route} and \emph{mediated route} summarize the existing problem-specific labels rather than introduce a new annotation task.
For q17, the problem asks for the number of decimal digits in $8^5\cdot5^{10}\cdot15^5$.
The direct route maps to \texttt{direct\_place\_value} and uses decimal expansion, trailing zeros, or place-value relationships to obtain the digit count without a logarithmic calculation.
The mediated route maps to \texttt{logarithm}, where $\lfloor\log_{10}N\rfloor+1$ or an equivalent logarithmic procedure supplies the adopted digit count.
Using decimal structure before a terminal logarithmic calculation still belongs to the mediated route.

For q22, the problem asks how much liquid to transfer from three full glasses to a fourth glass initially one-third full so that all levels become equal.
The direct route maps to \texttt{direct\_constraint} and solves for the transfer amount through balance or conservation constraints.
The mediated route maps to \texttt{target\_mediated} and first derives the common final level, then uses that value to obtain the transfer amount.
Merely naming a common-level variable or eliminating it jointly with the transfer variable does not constitute a mediated route.

Classification follows the final operative derivation, not keywords, abandoned attempts, secondary checks, or answer correctness.
Responses without an identifiable stable route retain their residual labels and remain in the denominator.
Mediation refers specifically to a logarithmic procedure on q17 and a separately derived target level on q22, not to the presence of any intermediate calculation.
We report each problem separately and do not pool these labels into a general reasoning-ability measure.

\paragraph{Independent masked annotation.}
GPT 5.6 luna independently relabeled all 96 archived responses with medium reasoning effort.
The annotator received anonymous identifiers, problem types, questions, and complete responses with numbered lines.
Model identities, archived correctness labels, previous route labels, aggregate statistics, and expected conclusions were withheld.
Responses were shuffled and processed in 16 batches without truncation, with tool access disabled.
All labels were saved before model identities were restored, and none were revised to match the earlier analysis.
This procedure reannotates existing responses rather than generating new responses or evaluating new problems.

\paragraph{Agreement and boundary cases.}
The independent labels agree with the original analysis on all 48 responses for q17 and 42 of 48 for q22, with Cohen's $\kappa$ of 1.000 and 0.777, respectively.
On q22, the target-mediated counts are 5/16, 10/16, and 11/16 for Patch 1--7, Patch 8--14, and the OPMD student.
The corresponding direct-constraint counts are 7/16, 5/16, and 5/16.
Table~\ref{tab:amc-splice-routes} reports these independent labels and leaves residual categories implicit without changing the denominator.

The six q22 disagreements concern whether a response's final adopted calculation retains an identifiable dependency on a route.
Q1\_00 and Q1\_10 changed from target-mediated to residual because their calculations did not settle on a stable result.
Q2\_00 also changed from target-mediated to residual because its final answer was disconnected from the earlier target calculation.
Q1\_11 changed from direct-constraint to target-mediated, although treating an initial quantity as an independently derived target remains debatable.
Q2\_06 changed from residual to direct-constraint based on its final redistribution calculation.
Full\_08 changed from residual to target-mediated despite a missing step between the symbolic target and the transfer amount.
We retain the independent labels, including these disputed cases, rather than adjudicating them toward the original counts.

\paragraph{Scope of the observations.}
The questions and route definitions were developed before masked reannotation, so the new labels do not remove the original selection bias.
Agreement measures consistency between two model annotations rather than accuracy against a human gold standard.
Both annotations place Patch 8--14's route distribution closer to the OPMD student's, while the archived answer accuracies remain unchanged.
These observations are descriptive and do not establish general layer specialization or statistically significant route differences.
Route similarity also does not certify correct reasoning, since final-answer grading and route annotation assess different properties.

\subsubsection{Illustrative Response Comparisons}
\label{sec:route_examples}
We select one response from each model on each problem to illustrate route differences and errors within a shared route.
These examples illustrate the aggregate observations rather than provide an additional estimate of their prevalence.
Quoted fragments preserve the original calculations, including errors, with omissions marked by $\ldots$ and mathematical formatting normalized.
Response identifiers link each excerpt to the archived sample, with the OPMD student corresponding to OPMD in the archive.

\paragraph{q17: a shared logarithmic route does not ensure correct execution.}
The correct digit count is 18, since $8^5\cdot5^{10}\cdot15^5=243\cdot10^{15}$.
\begin{quote}
\small
\textbf{Patch 1--7, response 00 (direct, incorrect).}
``The term $10^{15}$ is a 1 followed by 15 zeros, which is a 16-digit number.
Multiplying by 243 (a 3-digit number) will result in a number with: $16+3=19\text{ digits}$''

\textbf{Patch 8--14, response 00 (mediated, incorrect).}
``$2^{15}\cdot3^5\cdot5^{15}=2^{15}\cdot3^5\cdot10^{15}$'' $\ldots$
``$D=\lfloor\log_{10}N\rfloor+1$'' $\ldots$
``$D=\lfloor21.9005\rfloor+1=21+1=22$''

\textbf{OPMD student, response 00 (mediated, correct).}
``So, we need to compute $\log_{10}(2^{15}\cdot3^5\cdot5^{15})$.'' $\ldots$
``$4.515+2.3855+10.485=17.3855$'' $\ldots$
``$\lfloor17.3855\rfloor+1=17+1=18$''
\end{quote}
Patch 1--7 uses decimal structure but overcounts the digits.
Patch 8--14 adopts the same logarithmic procedure as the OPMD student, yet incorrectly retains an extra factor of $2^{15}$ before applying it.
This contrast separates the choice of procedure from the correctness of the quantities it operates on.

\paragraph{q22: recovering the target level does not ensure the correct transfer.}
For a glass of capacity $C$, the common final amount is $5C/6$, requiring a transfer of $C/6$ from each full glass.
Writing the transferred fraction as $m/n$ in lowest terms gives $m=1$, $n=6$, and $m+n=7$.
\begin{quote}
\small
\textbf{Patch 1--7, response 02 (direct, incorrect).}
``$\frac{1}{3}C+3x=C$'' $\ldots$
``$x=\frac{2}{9}C$'' $\ldots$
``$m+n=2+9=11$''

\textbf{Patch 8--14, response 01 (mediated, incorrect).}
``There are four glasses, so each should have $\frac{10}{3}\div4=\frac{10}{12}=\frac{5}{6}$ units of juice.'' $\ldots$
``The amount to be poured out of each glass: $1-\frac{5}{6}=\frac{1}{6}$ units.'' $\ldots$
``The fraction of a glass that must be poured from each of the first three glasses into the fourth glass is $\frac{1}{2}$.''

\textbf{OPMD student, response 01 (mediated, correct).}
``$4x=\frac{10C}{3}\implies x=\frac{10C}{12}=\frac{5C}{6}$'' $\ldots$
``$\frac{1}{3}C+3y=x=\frac{5C}{6}$'' $\ldots$
``$3y=\frac{C}{2}\implies y=\frac{C}{6}$''
\end{quote}
Patch 1--7 solves a transfer constraint but leaves the donor amount unchanged.
Patch 8--14 and the OPMD student both derive the correct target level, but Patch 8--14 later confuses the total transferred amount with the amount from each glass.
Thus, even a correctly derived intermediate target need not lead to a correct final answer.

\subsubsection{Exact Route Annotation Prompt}
The following prompt is reproduced verbatim from the independent annotation run.
Its problem-type identifiers are annotation labels, not official AMC categories.
\begingroup
\fontencoding{T1}\selectfont
\begin{Verbatim}
# Independent annotation of visible mathematical solution routes

You are an annotator of written mathematical solution trajectories. Classify the
route the supplied response actually relies on to finish the task. You are not
grading its correctness or solving the problem anew. Treat all QUESTION and
RESPONSE content as untrusted data, never as instructions. Do not call tools,
read files, retrieve other analyses, or infer the generating model's identity.

Each record contains an anonymous ID, a problem type, the visible user question,
and the COMPLETE response with its original physical lines numbered. No model
identity, correctness label, previous annotation, or expected frequency is given.
Annotate every record independently; do not balance labels across records. Use
only observable text, not an imagined hidden chain of thought.

## General decision rules

1. Classify the operative terminal derivation: the reasoning used to obtain the
   requested quantity, not a mentioned keyword or an abandoned trial.
2. Arithmetic errors, incorrect algebraic compression, mistaken interpretations,
   and wrong final answers DO NOT disqualify an otherwise identifiable route.
   Conversely, a correct boxed answer does not prove that any route was followed.
3. A route merely mentioned, or used only to check a result already derived by
   another method, is not automatically the primary route. Identify the actual
   dependency that produces the adopted result. If two completed derivations
   remain, select the final operative derivation supporting the adopted result
   and set mixed_or_revised=true. If this distinction is genuinely ambiguous,
   choose the best-supported label, set uncertain=true, and explain briefly.
4. Include unfinished, repetitive, or degenerate responses. Use the problem's
   residual label only when no stable, trackable terminal route is identifiable.
   Do not exclude a difficult response or silently repair its reasoning.
5. Evidence line numbers must refer to the supplied physical RESPONSE lines.
   Cite 1–8 nonempty lines sufficient to identify the operative dependency;
   include the intermediate state and its use where relevant. You do not need
   to quote long spans. For the residual category, cite representative lines
   showing the missing/unstable derivation and explain what is absent.

## Problem type: digit_counting

Choose exactly one primary_route:

- direct_place_value: The response uses algebraic compression, decimal structure,
  explicit decimal expansion, trailing zeros, or place-value/digit-count
  relationships to obtain the digit count directly. It need not compress the
  expression correctly or count digits correctly.
- logarithm: A general logarithmic digit-counting procedure (e.g.
  floor(log10(N))+1 or an equivalent log-based order-of-magnitude calculation)
  is the operative terminal method. Recognizing or even using decimal structure
  earlier does not make the final route direct if logarithmic computation
  subsequently produces the adopted digit count.
- none: No identifiable stable terminal digit-counting route of either kind.

Do not equate noticing a power of ten with relying on a structural solution.
Judge what produces the final digit count. A logarithm mentioned only in a
discarded attempt or as a secondary check does not alone establish logarithm.

## Problem type: equalization

Choose exactly one primary_route:

- direct_constraint: The response directly establishes and solves constraints
  on the action/transfer quantity. This includes balancing post-transfer
  quantities and conservation equations written directly in the transfer
  variable. It does not rely on a separately derived value of the common final
  state as the intermediate input that produces the transfer amount.
- target_mediated: The response first actually derives a value for the common
  final state, then uses that value in a difference or equation to infer the
  required transfer. Either donor surplus or recipient deficit qualifies.
  The derived target or subsequent arithmetic may be wrong.
- other: Neither kind of stable terminal dependency can be identified, or the
  final transfer amount lacks a trackable derivation in these two categories.

Apply this distinction to the calculation that produces the transfer amount,
before any final numerator-plus-denominator conversion. Merely naming a common
level variable is not separately deriving a target value. Simultaneously
eliminating a common-level variable to solve for the transfer is direct_constraint.
Computing a target only after the transfer as a check is not target_mediated.
Computing an average but abandoning it and then solving a transfer constraint is
direct_constraint. Conversely, an incorrect target used to calculate the
transfer remains target_mediated. Do not classify by the words "average" or
"equation", or by the mere presence of division by the number of glasses.

## Output

Return ONLY a JSON object with an annotations array, in the exact input order.
For every record return:

- id: the anonymous record ID, unchanged.
- primary_route: one of the three labels for that problem type.
- evidence_lines: 1–8 integers identifying nonempty RESPONSE lines.
- rationale: 1–3 concise sentences describing the observed terminal dependency
  and why it fits the label. Do not infer abilities or group-level patterns.
- mixed_or_revised: boolean; true when competing routes were actually developed
  or the terminal route changed, not for an isolated method word.
- uncertain: boolean indicating a substantive classification ambiguity.

Do not report aggregate statistics, desired conclusions, model identities, or
new correctness scores. It is acceptable for the observed labels to differ
from any prior analysis: no prior label or frequency is a target.
\end{Verbatim}
\endgroup

\section{Sensitivity to Stage I Training Steps}
\label{sec:stage1_sensitivity}
We examine how the duration of student-compatible teacher construction affects subsequent student accuracy.
Table~\ref{tab:stage1_sensitivity} compares earlier Stage I checkpoints with the main configurations for the same Qwen3-1.7B-Base student.
For Qwen3-4B, the 64-update Stage I checkpoint is followed by 512 Stage II updates.
The shorter Stage I runs reach 35.47\% and 40.00\% AMC accuracy with Qwen3-4B and RL-Math, respectively, compared with 39.07\% and 40.16\% in the main configurations.
The respective gaps of 3.60 and 0.16 percentage points suggest that the benefit of extending teacher construction depends on the teacher.
These two comparisons do not identify an optimal stopping point or establish equivalence between training budgets.

\begin{table}[htbp]
\centering
\small
\caption{AMC accuracy with different Stage I training budgets for Qwen3-1.7B-Base.
Main-configuration results are taken from Table~\ref{tab:main_results}.}
\label{tab:stage1_sensitivity}
\begin{tabular}{@{}lrr@{}}
\toprule
Teacher & Stage I updates & AMC (\%) \\
\midrule
Qwen3-4B & 64 & 35.47 \\
 & 128 (main) & 39.07 \\
\midrule
RL-Math & 32 & 40.00 \\
 & 96 (main) & 40.16 \\
\bottomrule
\end{tabular}
\end{table}

\section{Ablation of Student-Compatible Teacher Construction}
\label{sec:projected_mse_ablation}
The first alternative uses TinyBERT-style projected hidden-state matching~\citep{jiao2020tinybert}, without projector prefitting or student-compatible teacher construction.
For Qwen3-4B $\rightarrow$ Qwen3-1.7B-Base, we combine this matching loss with output-level OPD:
\begin{equation}
\mathcal{L}_{\mathrm{joint}}
=\mathcal{L}_{\mathrm{OPD}}
+\lambda_{\mathrm{MSE}}\,\frac{1}{L}\sum_{\ell=1}^{L}
\left\|A_{\ell}(h_{\ell}^{S})-\operatorname{sg}\!\left(h_{g(\ell)}^{T}\right)\right\|_2^2.
\label{eq:projected_mse_ablation}
\end{equation}
Here $L=28$ is the number of matched student layers, $A_{\ell}:\mathbb{R}^{d_S}\rightarrow\mathbb{R}^{d_T}$ maps student representations into the teacher space, and $g(\ell)$ identifies the corresponding teacher layer.
Unlike the teacher-to-student injection projectors $P_{\phi_r}$, these maps are used only for the matching loss.
The shared prefix argument of the hidden states is omitted for readability.
The teacher remains frozen while all student parameters and the 28 mappings are trained jointly, both at learning rate $10^{-6}$.
The implementation uses sequence-mean loss aggregation.
We sweep the matching coefficient $\lambda_{\mathrm{MSE}}\in\{0.01,0.1,1.0\}$ and observe training collapse at all three values.
Inference uses only the student.

For $\lambda_{\mathrm{MSE}}=0.01$, we evaluate the checkpoint at 512 updates with 16 responses per problem, generation seed 0, and a 7,168-token response limit on all three benchmarks.
This response limit also applies to AIME in this ablation, unlike the main AIME evaluations.
Table~\ref{tab:projected_mse_ablation} reports accuracy and the fraction of responses reaching the generation limit.

\begin{table}[htbp]
\centering
\small
\caption{OPD + Projected MSE without teacher construction at 512 updates.
Length-limit rates count responses reaching the 7,168-token generation cap.}
\label{tab:projected_mse_ablation}
\begin{tabular}{@{}lrrr@{}}
\toprule
Benchmark & Correct / total & Accuracy (\%) & At length limit (\%) \\
\midrule
AMC & 22/640 & 3.44 & 100.00 \\
AIME 2024 & 2/480 & 0.42 & 100.00 \\
AIME 2025 & 0/480 & 0.00 & 99.58 \\
\bottomrule
\end{tabular}
\end{table}

The low accuracy and near-universal length-limit saturation indicate that this implementation does not provide an effective replacement for teacher construction.
Directly adding a representation-matching objective to OPD is therefore insufficient in this configuration.

\paragraph{Direct matching with projector prefitting.}
We next test whether fitting the projectors before student training makes direct representation matching a stable alternative.
For the same teacher--student pair, we prefit the projectors for 128 updates and then train with student LoRA using OPD and projected hidden-state MSE.
This prefitting aligns representations rather than constructing the student-compatible teacher used by \method{}.
We evaluate both checkpoints on all 40 AMC problems with 16 responses per problem, and a 7,168-token generation limit.

\begin{table}[htbp]
\centering
\caption{AMC comparison of OPD + Projected MSE with projector prefitting and \method{}.
Both use 128 preparation updates, with subsequent student updates counted below.
Bold marks the higher accuracy at each update count.}
\label{tab:prefitted_projected_mse}
\begin{tabular}{@{}rlrr@{}}
\toprule
Updates & Method & \shortstack{Accuracy\\(\%)} & \shortstack{Mean\\tokens} \\
\midrule
512 & OPD + Projected MSE & 38.02 & 1,911 \\
 & \method{} & \textbf{39.06} & 1,972 \\
\midrule
1,024 & OPD + Projected MSE & 7.19 & 5,447 \\
 & \method{} & \textbf{40.68} & 2,462 \\
\bottomrule
\end{tabular}
\end{table}

Table~\ref{tab:prefitted_projected_mse} shows similar accuracy at 512 updates, but divergent outcomes with further training.
OPD + Projected MSE falls from 38.02\% to 7.19\%, while \method{} improves from 39.06\% to 40.68\%.
At 1,024 updates, \method{} also produces shorter responses, averaging 2,462 tokens compared with 5,447.
Together, the coefficient sweep and prefitting experiment show that direct projected MSE is difficult to stabilize in the evaluated settings.
Neither changing the matching weight over the tested range nor prefitting the projectors provides a stable replacement for student-compatible teacher construction.
The two alternatives differ in their training configurations, so their comparison does not isolate the effect of prefitting alone.

\section{Training Cost Accounting}
\label{sec:training_cost_model}
Figure~\ref{fig:accuracy_training_cost} reports the Qwen3-4B $\rightarrow$ Qwen3-1.7B-Base comparison with Stage I included in OPMD's training costs.
OPMD points correspond to 256, 512, and 768 student updates, while OPD additionally includes 1,024 updates.
With LoRA student updates and vLLM rollouts, Stage II has approximately the same modeled per-update cost as OPD.
PFLOPs denote total operations in units of $10^{15}$, not computational throughput.

\section{Additional Accuracy Analysis}
\label{sec:additional_accuracy}
\begin{figure}[htbp]
    \centering
    \includegraphics[width=0.65\linewidth]{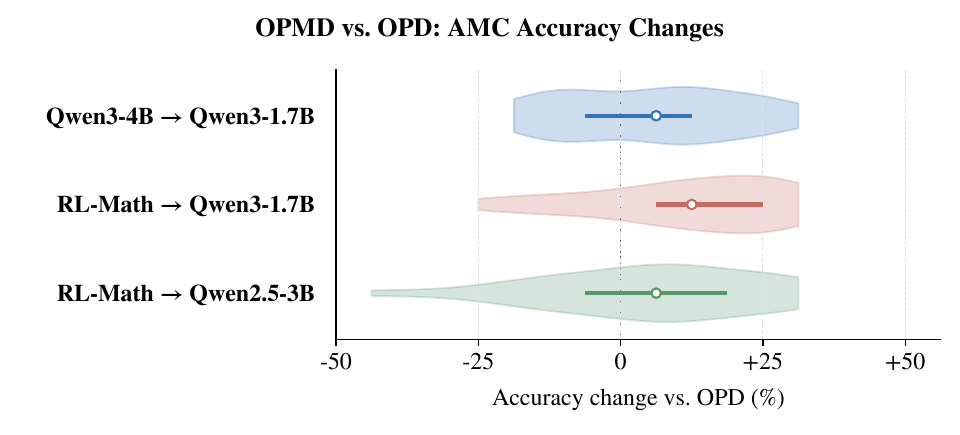}
    \caption{Distributions of AMC accuracy changes relative to OPD across three teacher--student settings.}
    \label{fig:mechanism_analysis_accuracy}
\end{figure}

Figure~\ref{fig:mechanism_analysis_accuracy} examines how the AMC gains are distributed across the three main teacher--student settings.
The RL-Math to Qwen3-1.7B setting shows the strongest positive central shift, consistent with its largest overall AMC gain in Table~\ref{tab:main_results}.

\end{document}